\PassOptionsToPackage{table,xcdraw}{xcolor}
\documentclass[runningheads]{llncs}

\usepackage{graphicx}
\usepackage{booktabs}
\usepackage{multirow}
\usepackage{array}
\usepackage{tabularx}
\usepackage{adjustbox}
\usepackage{colortbl}
\usepackage{makecell}
\usepackage{enumitem}
\usepackage{threeparttable}
\usepackage{tablefootnote}
\usepackage{geometry}
\usepackage{color} 
\usepackage{xspace}
\usepackage{amsmath}
\usepackage{balance}
\usepackage{subfig}
\usepackage{cite}
\usepackage{comment}
\usepackage{orcidlink}
\usepackage{hyperref}
\usepackage{longtable}

\definecolor{headercolor}{rgb}{0.9, 0.9, 0.9}

\renewcommand{\arraystretch}{1.5} 

\newcommand{\cs}{\textsc{ClickSight}\xspace}
\newcommand{\compl}{\textit{Completeness}\xspace}
\newcommand{\corr}{\textit{Correctness}\xspace}
\newcommand{\just}{\textit{Justifiedness}\xspace}
\newcommand{\compr}{\textit{Comprehensibility}\xspace}

\newcommand{\zs}{\textit{Zero-shot}\xspace}
\newcommand{\ct}{\textit{Chain-of-Thought}\xspace}
\newcommand{\meta}{\textit{Meta-Prompting}\xspace}
\newcommand{\cop}{\textit{Chain-of-Prompts}\xspace}
\newcommand{\ps}{PharmaSim\xspace}
\newcommand{\bl}{Beer's Law Lab\xspace}
\usepackage{balance}
\begin{document}
\title{
ClickSight: Interpreting Student Clickstreams to Reveal Insights on Learning Strategies via LLMs}
%
%
%
%
%
\author{
Bahar Radmehr~\orcidlink{0009-0008-8034-4853}\inst{1} \and
Ekaterina Shved~\orcidlink{0009-0004-8915-3640}\inst{1} \and
Fatma Betül Güreş~\orcidlink{0000-0002-8664-1022}\inst{1} \and \\
Adish Singla~\orcidlink{0000-0001-9922-0668}\inst{2} \and
Tanja Käser~\orcidlink{0000-0003-0672-0415}\inst{1}}

\authorrunning{B. Radmehr et al.}
%

\institute{
EPFL, Switzerland \\
\email{\{bahar.radmehr, kate.kutsenok, fatma-betul.gures, tanja.kaeser\}@epfl.ch}
\and
MPI-SWS, Germany \\
\email{adishs@mpi-sws.org}
}
\maketitle              
\vspace{-5mm}
\begin{abstract}
Clickstream data from digital learning environments offer valuable insights into students' learning behaviors, but are challenging to interpret due to their high dimensionality and granularity.
Prior approaches have relied mainly on handcrafted features, expert labeling, clustering, or supervised models, therefore often lacking generalizability and scalability. 
In this work, we introduce \cs, an in-context Large Language Model (LLM)-based pipeline that interprets student clickstreams to reveal their learning strategies. \cs takes raw clickstreams and a list of learning strategies as input and generates textual interpretations of students' behaviors during interaction. 
We evaluate four different prompting strategies and investigate the impact of self-refinement on interpretation quality. Our evaluation spans two open-ended learning environments and uses a rubric-based domain-expert evaluation. 
Results show that while LLMs can reasonably interpret learning strategies from clickstreams, interpretation quality varies by prompting strategy, and self-refinement offers limited improvement. 
\cs demonstrates the potential of LLMs to generate theory-driven insights from educational interaction data.
\keywords{learning strategies  \and student clickstreams \and large language models \and log data \and open-ended learning environments}
\end{abstract}

\section{Introduction}
\label{sec:intro}

The rise of digital learning environments has enabled more personalized, scalable, and data-driven educational experiences~\cite{moore2013interactive}. These environments facilitate gathering rich clickstream data---detailed, time-stamped records of student interactions---which can offer valuable insights into learning behaviors. Such insights can assist researchers in understanding students' behaviors better, help instructors provide timely interventions to struggling students, and assist students in improving their learning strategies by providing feedback\cite{click1}. However, extracting meaning from clickstreams remains challenging due to their high dimensionality and granularity that makes expert annotation labor-intensive and cognitively demanding\cite{clickstream-difficulty}.

\looseness-1Prior work in AI for education has employed a range of techniques to detect learning strategies and analyze student behavior from clickstreams. These include designing domain-specific features for various learning strategies~\cite{bll,DBLP:journals/aiedu/KaserS20}, clustering students based on behavioral patterns~\cite{bll}, sequential pattern mining to uncover frequent interaction traces~\cite{edm2020_classifier_pattern,aied2024_seq_dashboard,DBLP:journals/aiedu/ZhangBH22}, supervised classifiers trained on annotated data~\cite{edm2020_classifier_pattern}, and knowledge tracing to model evolving knowledge states~\cite{DBLP:journals/aiedu/KaserS20}. While these methods yield valuable insights, they often require significant manual effort and domain expertise and are difficult to generalize across contexts.

Large Language Models (LLMs) like GPT-4o~\cite{openai2024gpt4osystemcard} offer new opportunities to support educational technology~\cite{DBLP:journals/corr/abs-2402-01580}. LLMs have been applied to a range of educational tasks, including content generation~\cite{aied2024_gpt_generation}, feedback provision~\cite{lak_feedback}, and student simulation~\cite{edm_student_modeling}. Given their strong ability to recognize patterns~\cite{Mirchandani2023LargeLM}, LLMs hold promise for interpreting clickstream data through the lens of learning strategies. However, this potential remains largely unexplored.


In this paper, we introduce \cs, an in-context LLM-based pipeline for interpreting student clickstreams via predefined learning strategies. Given a student's clickstream and a list of learning strategies, \cs interprets the student’s behavior. We consider four prompting strategies and examine the impact of self-refinement, where the LLM critiques and revises its own outputs.
We extensively evaluate \cs across two environments from different domains: \textbf{\ps}~\cite{ps}, a simulation for pharmacy assistant diagnostic training, and \textbf{\bl}~\cite{bll}, a virtual chemistry lab for secondary level students. To assess interpretation quality, we develop a rubric with four criteria grounded in human-centered explanation theory, and conduct domain expert evaluation based on it. We address two research questions: (1) which prompting strategy leads to the best strategy-based interpretation of students' clickstreams \textbf{(RQ1)}, and (2) how does self-refinement affect interpretation quality \textbf{(RQ2)}?

Our results show that \cs enables high-quality interpretations, with one prompting strategy outperforming others. Self-refinement leads to modest improvements for some strategies, while introducing errors in others. Our implementation of \cs is available on our GitHub repository: \href{https://github.com/epfl-ml4ed/ClickSightLBR25}{https://github.com/epfl-ml4ed/ClickSightLBR25}.

\vspace{-4mm}
\section{Methodology}\label{sec:method}
\vspace{-3mm}

\begin{figure}[t]
    \centering
    \includegraphics[width=0.86\linewidth]{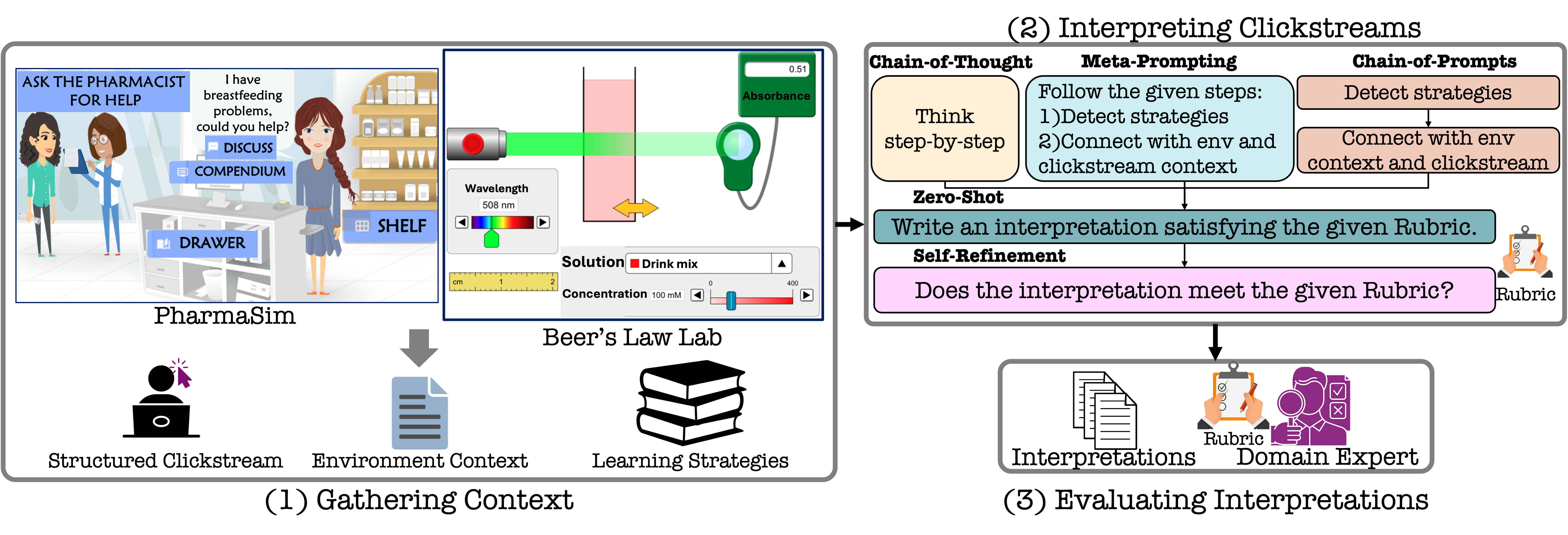}
\caption{The \cs pipeline interprets student clickstreams in three stages. (1) clickstream data from \ps and \bl are structured, and context and learning strategies are gathered; (2) an LLM is prompted using one of four prompting strategies (\zs, \ct, \meta, \cop) with optional self-refinement to generate interpretations; (3) human experts assess the outputs using a quality rubric.}
    \label{fig:pipeline}
    \vspace{-4mm}
\end{figure}

\looseness-1 \cs consists of three stages (see Fig.\ref{fig:pipeline}). In the first stage, we describe the simulation setting, identify relevant learning strategies, and organize student clickstreams into a structured format. The next stage prompts an LLM to generate textual interpretations grounded in the contextual information constructed in the first stage. Finally, we apply a rubric-based domain-expert evaluation to assess the quality, clarity, and relevance of the generated interpretations.

\vspace{-3mm}
\subsection{Gathering Environment Context}\label{sec:env-context}
\vspace{-1mm}
We use data from two different educational environments \ps~\cite{ps} and \bl~\cite{bll}.

\vspace{1mm} \noindent  \textbf{\ps} 
is a scenario-based learning environment for pharmacy assistant training, where students develop diagnostic skills through simulated interactions with a virtual patient. In the selected scenario (see Fig.~\ref{fig:pipeline}), a mother visits a pharmacy for a breastfeeding issue, which the student investigates by performing actions such as asking the patient questions, seeking help from the pharmacist, researching medicines, examining shelf products, and consulting medical documentation.

\vspace{1pt} \noindent \textit{Student Clickstreams for \ps.}  
Following the representations used in existing literature~\cite{Radmehr_Singla_Käser_2025}, we structured each clickstream as a sequence of function calls corresponding to specific action types in the simulation. For example, the action of asking the virtual mother about her symptoms at time $t$ is represented as:
\texttt{discuss(mother, symptoms, $t$) [output: My breast hurts...]}.

\vspace{1pt} \noindent \textit{Learning Strategies for \ps.}
\looseness-1We gathered \textbf{nine} diagnostic strategies observed in student behavior in environments like \ps, based on prior work. These vary in alignment with learning goals and span different aspects of the environment.
Among diagnostic conversation strategies highlighted by prior work~\cite{q2,q3}, effective strategies include \textit{LINDAAFF}, where students ask targeted questions about symptom details, and \textit{Inquiry about Relevant External Factors}, which addresses influences like the baby’s health. Ineffective approaches include \textit{Premature Closure} (settling on a diagnosis too early), \textit{Random Inquiry} (asking about irrelevant topics), and \textit{Insufficient Inquiry} (failing to collect enough information).
\textit{Research strategy} describes how students use the documentation, compendium, and shelf. Inspired by prior work on self-regulation~\cite{zimmerman2002becoming} this strategy includes three types: Targeted Research, where students focus on relevant materials; Unfocused Research, characterized by scattered, aimless browsing; and Minimal Research, involving little or no use of these tools.
\textit{Hint-seeking strategy} can take the form of Thoughtful Hint Seeking, where students request hints at appropriate moments after meaningful engagement; Premature Hint Seeking, in which students rely on hints too early without sufficient exploration; and No Hint Seeking, where students do not request any hints during the interaction~\cite{help}.
Finally, students may engage in \textit{Iterative Reflection}, pausing between key actions to process and consolidate information~\cite{pause}, or exhibit \textit{Gaming the System}, where they rush through tasks without meaningful engagement~\cite{baker2006gaming}.

\vspace{1mm} \noindent \textbf{\bl} allows students to investigate how cuvette width, solution concentration, and laser color-solution combination affect light absorbance (see Fig.~\ref{fig:pipeline}). Width and concentration have a linear relationship with absorbance, while laser-solution color follows a non-linear pattern.

\vspace{1pt} \noindent \textit{Student Clickstreams for \bl.}  
Students engage in two action types: \textit{exploration}, manipulating variables and observing changes in absorbance, and \textit{analysis}, recording values for later investigation. Since our focus is on strategy-based interpretations, we analyze only exploration actions, which are formatted as function calls, such as: \texttt{explore(variable=width, value\_changes=increase from 1.0cm to 1.2cm, begins=5:55, duration=0.2s) [absorbance: increase from 0.67 to 0.79]}, with consecutive actions on the same variable within $3$ seconds merged following~\cite{bll}.

\vspace{1pt} \noindent \textit{Learning Strategies for \bl.}  
We analyzed student behaviors using three core inquiry strategies from prior work~\cite{bll}. The \textit{Control of Variables Strategy (CVS)} involves manipulating one independent variable at a time while holding others constant. The \textit{Range} strategy encourages exploring a full span of values to identify trends. The \textit{Optimal} strategy focuses on selecting values for non-focal variables that minimize confounding effects (e.g., avoiding  laser-solution color match when testing cuvette width). These strategies vary between the three independent variables and over time, making them more difficult to interpret than the aggregated strategies in \ps.

\vspace{-3mm}
\subsection{Interpreting Clickstreams}\label{sec:self-refinement}
\vspace{-1mm}
At the core of \cs is an LLM that generates interpretations of student clickstreams using the learning strategies and context gathered in stage 1. The LLM is guided by a rubric based on human-centered explanation research. We explored four prompting strategies: \zs, \ct, \meta, and \cop, and additionally included a self-refinement step that allows the LLM to review and improve its responses.

\begin{table}[t]
    \caption{Quality evaluation rubric for clickstream interpretation.} 
    \label{tab:rubric}
    \centering
    \begin{adjustbox}{width=\textwidth}
    \renewcommand{\arraystretch}{0.95}
    \begin{tabular}{lp{15cm}}
\toprule
\textbf{Criterion} & \textbf{Definition} \\
\midrule
\textbf{Completeness} & 
Percentage execution level of learning strategies across different aspects mentioned in the interpretation.\\

\textbf{Correctness} & 
Percentage of correct execution levels of strategies stated in the interpretation based on the clickstream. \\

\textbf{Justifiedness} & 
Percentage of conclusions about strategy execution levels linked to evidence in the clickstream. \\


\textbf{Comprehensibility} & 
Whether the explanation is clear and easy to understand, without ambiguity.\\

\bottomrule
\end{tabular}
\end{adjustbox}
\vspace{-4mm}
\end{table}


\vspace{1mm} \noindent \textbf{Rubric for High-Quality Interpretations.}
To guide the LLM in producing meaningful interpretations, we included a detailed description of what constitutes a high-quality interpretation directly in the prompt. Based on prior work on human-centered explanation quality~\cite{rubric1,rubric3}, the rubric outlines four criteria (see Table~\ref{tab:rubric}): \compl, \corr, \just, and \compr. \compl ensures that all levels of execution of learning strategies in different aspects of the environment are included. \corr requires that stated execution levels align with the actual clickstream data and are unmet for unmentioned strategies. \just requires that conclusions are supported by evidence from the clickstream (e.g., student actions). \compr ensures clarity, and is considered unmet if the interpretation contains any ambiguity. This rubric was used to shape LLM responses and assess the quality of the generated interpretations.

\vspace{1mm} \noindent \textbf{Prompting Strategies.}
To examine how prompting affects interpretation quality, we designed four strategies: \zs, \ct, \meta, and \cop, drawing on prior work in effective LLM prompting. Each prompt includes the student’s clickstream, environment context, learning strategies, and the rubric. The detailed prompts are available on our GitHub.
\begin{itemize}
    \item \noindent\zs. The LLM is instructed to generate an interpretation satisfying the rubric, without giving any reasoning guidance, relying entirely on its internal capabilities about the task~\cite{DBLP:journals/corr/abs-2205-11916zeroshot}.

    \item \noindent\ct. Adds a segment encouraging step-by-step reasoning, prompting the LLM to “think step-by-step” while building the interpretation incrementally~\cite{DBLP:conf/nips/Wei0SBIXCLZ22cot}.

    \item \noindent\meta. Introduces multi-step reasoning instruction within a single prompt, instructing the LLM to detect strategy execution levels, link it to the environment context and student's actions, and generate a final interpretation based on prior steps~\cite{zhang2025metapromptingaisystems}.

    \item \noindent\cop. Uses three sequential prompts: (1) detect strategy execution levels, (2) connect to environment context and student's actions, and (3) generate a final interpretation~\cite{sun-etal-2024-promptchain}.
\end{itemize}

\noindent \textbf{Self-Refinement.}
To improve alignment with the rubric criteria, we introduce a self-refinement step that can follow any of the prompting strategies. After generating an initial interpretation, the LLM evaluates it using binary questions targeting the four rubric criteria. The first three, \compl, \corr, and \just, are each assessed with multiple questions, one per strategy within a specific environment aspect. \compr is evaluated with a single question. If any response is negative, the LLM generates targeted feedback and revises its interpretation. This process can be repeated; we use three rounds, which we found sufficient for meaningful refinement. We refer to the initial version as \textit{Initial} and the final version as \textit{Self-Refined}.


\vspace{-4mm}
\subsection{Evaluating Interpretations}\label{sec:eval}
\vspace{-2mm}
In this stage, human experts evaluate the generated interpretations using the rubric criteria. Two independent learning science experts per environment rated each output with a structured grading scheme reflecting how well it satisfies each rubric dimension.

\vspace{1mm} \noindent \textbf{Grading Scheme.}
Following the approach in~\ref{sec:self-refinement}, we used a series of yes/no questions to guide expert evaluation for each rubric criterion. The first three—\compl, \corr, and \textit{Justifiedness}—were each assessed using nine targeted questions, one per strategy-aspect pair: \ps includes $9$ strategies within one aspect, and Beer's Law Lab includes $3$ strategies across $3$ aspects. \compr was evaluated with a single binary question.

\vspace{1mm} \noindent \textbf{Grading Process.} To evaluate each behavior profile identified through clustering, we sampled five representative students per cluster. This resulted in 30 evaluation instances for \ps ($5 \times 6$ clusters derived from k-means clustering of clickstreams, with the optimal $k$ determined via the elbow method) and 20 for \bl ($5 \times 4$ clusters following the setup of prior work~\cite{bll}), resulting in 400 total interpretations ($50$ students $\times$ 4 prompting methods $\times$ 2 self-refinement conditions) to be annotated. All participants provided informed consent, and the studies were approved by the university's ethics committee (Nr. 010-2023 and 062-2021). To ensure consistency, two experts first independently evaluated interpretations of representative clickstreams from each environment, selecting one clickstream from each behavior cluster and interpreted it using all four prompting strategies without self-refinement, yielding 24 samples for \ps ($6$ clusters $\times$ $4$ methods) and 16 for Beer's Law Lab ($4$ clusters $\times$ $4$ methods). Inter-annotator agreement was measured using Cohen's $\kappa$ for each binary question~\cite{McHugh2012Kappa}. For multi-question criteria, we report both the average and minimum agreement.
For \ps, the agreement scores were: $\kappa_{\text{avg-complete}} = 1.0$, $\kappa_{\text{avg-correct}} = 0.95$ ($\kappa_{min} = 0.74$), $\kappa_{\text{avg-justified}} = 0.97$ ($\kappa_{min} = 0.86$), and $\kappa_{\text{comprehensible}} = 1.0$. 
For Beer's Law Lab, the agreement scores were: $\kappa_{\text{avg-complete}} = 1.0$ ($\kappa_{min} = 1.0$), $\kappa_{\text{avg-correct}} = 0.98$ ($\kappa_{min} = 0.89$), $\kappa_{\text{avg-justified}} = 0.91$ ($\kappa_{min} = 0.70$), and $\kappa_{\text{comprehensible}} = 1.0$.
Following high agreement, the remaining interpretations were annotated by a single expert.

\vspace{1mm} \noindent \textbf{Score Calculation.}  
To compute criterion-level scores, we averaged binary responses for multi-question criteria. \compr, assessed with a single yes/no question scored directly ($1$ for clear interpretations, $0$ for those with any ambiguity). To calculate an overall quality score ($0$–$1$) for each interpretation, we first gathered binary responses to the nine questions spanning the three multi-question criteria—\compl, \corr, and \just. For each strategy, a composite score was computed by multiplying its three criterion scores, ensuring its execution level was correctly mentioned and justified. We then averaged the nine strategy scores and multiplied this intermediate value by the \compr score, ensuring that overall quality guaranteed \compr as well.

\vspace{-3mm}
\section{Results}
\label{sec:results}
We evaluated \textsc{ClickSight} on clickstreams from \ps and \bl to assess the impact of prompting stratey (RQ1) and self-refinement (RQ2) on interpretation quality. 

\vspace{-3mm}
\subsection{Comparative Evaluation of Prompting Strategies (RQ1)}\label{sec:rq1}

We first compared the quality of interpretations generated by the four prompting strategies, \zs, \ct, \meta, and \cop, based on our evaluation rubric (Section~\ref{sec:eval}). Table~\ref{tab:prompting_comparison} reports the average scores (and standard deviations) for all criteria and environments.
\zs achieved the highest overall score in both environments, followed by \cop. It also attained a perfect score in \compl, while \ct scored lowest on this criterion. \ct’s intermediate reasoning steps often omitted some strategies, leading to incomplete interpretations.
In \corr, \meta outperformed other strategies in \ps, likely due to its detecting all strategies together in a single prompt, but this advantage did not hold in \bl. Both \ct and \meta scored lower in \just, often producing shorter outputs that lacked supporting evidence from the clickstream.
All strategies achieved perfect \compr in \ps, but only \zs maintained this in \bl. In other cases, outputs occasionally included raw, function-style clickstream elements that hindered readability, although all prompts instructed the LLM to describe student actions in natural language.

\begin{table}[t]
    \centering
    \renewcommand{\arraystretch}{1.3}
    \setlength{\tabcolsep}{2pt}
    \caption{Performance of different prompting strategies on \ps and \bl.}
    \label{tab:prompting_comparison}
    
    \begin{adjustbox}{max width=\textwidth}
    \begin{tabular}{>{\scriptsize}c>{\scriptsize}c>{\scriptsize}c>{\scriptsize}c>{\scriptsize}c>{\scriptsize}c|>{\scriptsize}c>{\scriptsize}c>{\scriptsize}c>{\scriptsize}c>{\scriptsize}c}
        \toprule
        & \multicolumn{5}{c|}{\textbf{\ps}} & \multicolumn{5}{c}{\textbf{\bl}} \\ 
        \multirow{-2}{*}{\makecell{\textbf{Prompting} \\ \textbf{Strategy}}} & \compl & \corr & \just & \compr & \textbf{Overall} & \compl & \corr & \just & \compr & \textbf{Overall} \\ 
        \midrule
        \textbf{\zs} & 
        \normalsize 1.00 {\scriptsize$\pm$0.00} & \normalsize 0.79 {\scriptsize$\pm$0.02} & \normalsize 1.00 {\scriptsize$\pm$0.00} & \normalsize 1.00 {\scriptsize$\pm$0.00} & \cellcolor[HTML]{C0C0C0}\normalsize \textbf{0.79 {\scriptsize$\pm$0.02}} & 
        \normalsize 1.00 {\scriptsize$\pm$0.00} & \normalsize 0.76 {\scriptsize$\pm$0.04} & \normalsize 1.00 {\scriptsize$\pm$0.00} & \normalsize 1.00 {\scriptsize$\pm$0.00} & \cellcolor[HTML]{C0C0C0}\normalsize \textbf{0.76 {\scriptsize$\pm$0.04}} \\ 
        
        \textbf{\ct} & 
        \normalsize 0.93 {\scriptsize$\pm$0.02} & \normalsize 0.80 {\scriptsize$\pm$0.03} & \normalsize 0.87 {\scriptsize$\pm$0.03} & \normalsize 1.00 {\scriptsize$\pm$0.00} & \cellcolor[HTML]{C0C0C0}\normalsize 0.66 {\scriptsize$\pm$0.04} & 
        \normalsize 0.87 {\scriptsize$\pm$0.05} & \normalsize 0.67 {\scriptsize$\pm$0.05} & \normalsize 0.19 {\scriptsize$\pm$0.07} & \normalsize 0.85 {\scriptsize$\pm$0.08} & \cellcolor[HTML]{C0C0C0}\normalsize 0.15 {\scriptsize$\pm$0.07} \\ 
        
        \textbf{\meta} & 
        \normalsize 0.96 {\scriptsize$\pm$0.01} & \normalsize 0.89 {\scriptsize$\pm$0.02} & \normalsize 0.74 {\scriptsize$\pm$0.04} & \normalsize 1.00 {\scriptsize$\pm$0.00} & \cellcolor[HTML]{C0C0C0}\normalsize 0.61 {\scriptsize$\pm$0.03} & 
        \normalsize 0.99 {\scriptsize$\pm$0.01} & \normalsize 0.72 {\scriptsize$\pm$0.04} & \normalsize 0.83 {\scriptsize$\pm$0.06} & \normalsize 0.90 {\scriptsize$\pm$0.07} & \cellcolor[HTML]{C0C0C0}\normalsize 0.56 {\scriptsize$\pm$0.06} \\ 
        
        \textbf{\cop} & 
        \normalsize 0.96 {\scriptsize$\pm$0.02} & \normalsize 0.77 {\scriptsize$\pm$0.02} & \normalsize 1.00 {\scriptsize$\pm$0.00} & \normalsize 1.00 {\scriptsize$\pm$0.00} & \cellcolor[HTML]{C0C0C0}\normalsize 0.74 {\scriptsize$\pm$0.03} & 
        \normalsize 0.99 {\scriptsize$\pm$0.01} & \normalsize 0.72 {\scriptsize$\pm$0.04} & \normalsize 0.98 {\scriptsize$\pm$0.01} & \normalsize 0.95 {\scriptsize$\pm$0.05} & \cellcolor[HTML]{C0C0C0}\normalsize 0.68 {\scriptsize$\pm$0.05} \\ 
        \bottomrule
    \end{tabular}
    \end{adjustbox}
\end{table}

\vspace{-3mm}
\subsection{Effect of Self-Refinement (RQ2)}
\begin{figure}[t]
    \centering
    \includegraphics[width=0.7\linewidth]{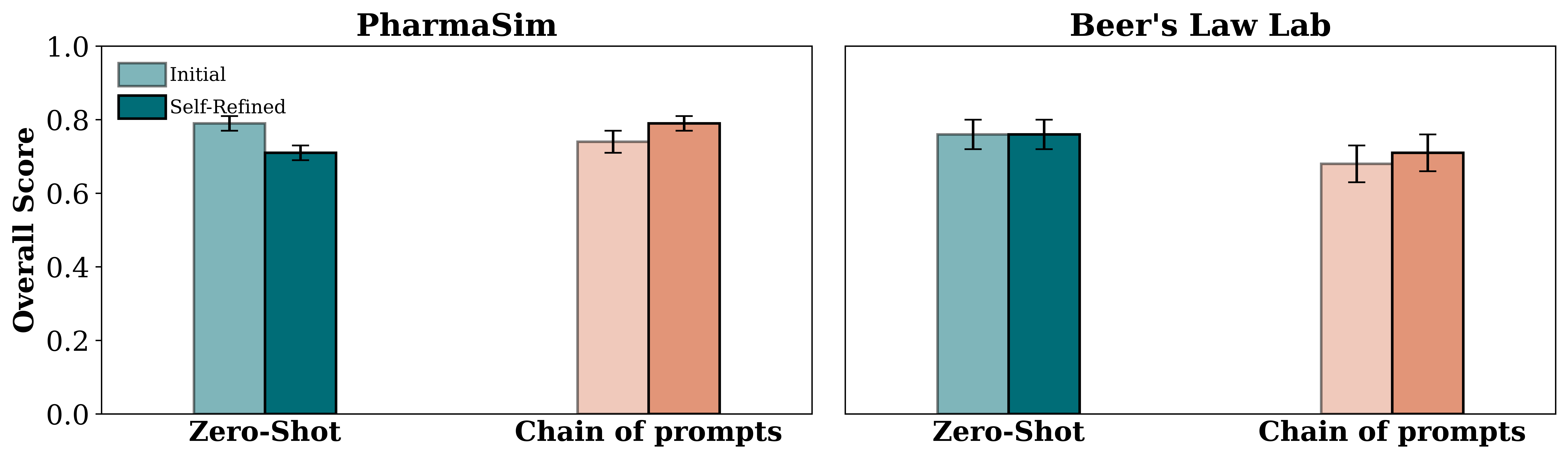}
\caption{Overall scores of \textit{Initial} and \textit{Self-Refined} interpretations generated using \zs and \cop prompting strategies in \ps and \bl.}
    \label{fig:self-refinement}
\end{figure}
In a second analysis, we examined the impact of self-refinement on interpretation quality. We applied self-refinement to the outputs from the two best-performing strategies, \zs and \cop. Figure~\ref{fig:self-refinement} shows the overall score for both \textit{Initial} and \textit{Self-Refined} interpretations. Refinement improved \cop, especially in \compl and \corr, across both environments. However, it slightly reduced \zs performance in \ps due to hallucinations during rubric self-evaluation. In \bl, it had no effect because the LLM assumed the initial output already satisfied all criteria.

\vspace{-3mm}
\section{Conclusion}
\label{sec:discussion}
\looseness-1In this paper, we introduced \cs, an in-context LLM framework for interpretation of student clickstreams, along with a rubric grounded in human-centered explanation theory. We evaluated four prompting strategies, with and without self-refinement, across two educational environments. \cs enabled LLMs to generate high-quality interpretations, with \zs performing best. Self-refinement showed mixed results, pointing to the need for refinement methods tailored to the task. While our study focused on zero-shot prompting, future work should explore few-shot settings and adapt outputs for stakeholders such as teachers and students. Overall, \cs demonstrates the potential of LLMs to generate theory-aligned insights from interaction data.

\vspace{1mm} \noindent\textbf{Acknowledgements.}
This project was substantially funded by the Swiss State Secretariat for Education, Research and
Innovation (SERI).


\bibliographystyle{splncs04}
\bibliography{bibliography}
\end{document}